\documentclass[runningheads]{llncs}
\usepackage{graphicx}
\usepackage{subcaption}
\usepackage{listings}
\usepackage{pgfplots,adjustbox}
\usepackage[title]{appendix}
\pgfplotsset{compat=1.14}
\usepackage[english]{babel}
\usepackage{makecell}
\usepackage{amsmath}

\usepackage[hyphens,spaces,obeyspaces]{url}
\urldef{\footurl}\url{https://www.europarl.europa.eu/RegData/etudes/STUD/2020/641530/EPRS_STU(2020)641530_EN.pdf}
\begin{document}
\title{Data Minimization for GDPR Compliance in Machine Learning Models}
\author{Abigail Goldsteen\inst{1} \and
Gilad Ezov\inst{1} \and
Ron Shmelkin\inst{1} \and
Micha Moffie\inst{1} \and
Ariel Farkash\inst{1} }
\authorrunning{A. Goldsteen et al.}
\institute{IBM Research - Haifa, Haifa University Campus, Haifa, Israel 
\email{\{abigailt,ronsh,moffie,arielf\}@il.ibm.com, Gilad.Ezov@ibm.com}\\
\url{http://www.research.ibm.com/labs/haifa/} }
\maketitle              % typeset the header of the contribution
\begin{abstract}
The EU General Data Protection Regulation (GDPR) mandates the principle of \emph{data minimization}, which requires that only data necessary to fulfill a certain purpose be collected. However, it can often be difficult to determine the minimal amount of data required, especially in complex machine learning models such as neural networks. We present a first-of-a-kind method to reduce the amount of personal data needed to perform predictions with a machine learning model, by removing or generalizing some of the input features. Our method makes use of the knowledge encoded within the model to produce a generalization that has little to no impact on its accuracy. This enables the creators and users of machine learning models to acheive data minimization, in a provable manner. 
\end{abstract}
\section{Introduction}
\label{sec:intro}
The European General Data Protection Regulation (GDPR)\footnote{https://ec.europa.eu/info/law/law-topic/data-protection/data-protection-eu\_en} dictates that ``Personal data shall be: adequate, relevant and limited to what is necessary in relation to the purposes for which they are processed''. This principle, known as \emph{data minimization}, requires that organizations and governments collect only data that is needed to achieve the purpose at hand. Organizations are expected to demonstrate that the data they collect is absolutely necessary, by showing concrete measures that were taken to minimize the amount of data used to serve a given purpose. Otherwise, they are at risk of violating privacy regulations, incurring large fines, and facing potential lawsuits. 

Advanced machine learning (ML) algorithms, such as neural networks, tend to consume large amounts of data to make a prediction or classification. Moreover, these ``black box'' models make it difficult to derive exactly which data influenced the decision. This type of algorithms is becoming prevalent in many business-related activities, including business intelligence, marketing, and sales. It is therefore increasingly difficult to show adherence to the data minimization principle. 

The recently published study of the European Parliamentary Research Service (EPRS) on the impact of the General Data Protection Regulation (GDPR) on artificial intelligence\footnote{\footurl} found that, although AI is not explicitly mentioned in the GPDR, many provisions in the GDPR are relevant to AI, specifically mentioning the principle of data minimisation. The authors propose that minimisation may require, in some contexts, reducing the 'personality' of the available data, rather than the amount of such data, i.e., it may require reducing, through technical measures, the ease with which the data can be connected to individuals.

A large amount of work is being done in the domain of privacy for  ML, spanning differential privacy, encryption, and more. We have a slightly different goal. In this work, we focus solely on accomplishing the target set out by the regulation, namely reducing the amount of personal data collected to only the data that can be demonstrated as necessary for fulfilling the purpose at hand. If the original purpose was to perform some analysis using an ML model, typically the model owner would not have any desire to reduce the accuracy of the model. Therefore, the goal of our method must be to retain, or get as close as possible to, the accuracy of the original model, while striving to reduce the amount of data collected. 

There may even be cases where all of the collected data is required to achieve the model's original accuracy; but one is still required to demonstrate that this is the case. However, as we will show in the following sections, in most cases the collected data can be reduced; and in these cases, we demonstrate that individuals' privacy is also improved.
Reducing the amount of data collected by organizations can bring additional benefits such as storage reduction, cost reduction, and decreased liability. 

We propose a method for data minimization that can reduce the amount and granularity of input data used to perform predictions by machine learning models. It is important to note that we are only concerned with minimizing newly collected data for analysis (i.e., runtime data), not the data used to train the model. Our method does not require retraining the model, and does not even assume the availability of the original training data. It therefore provides a simple and practical solution for addressing data minimization in existing systems. 

The type of data minimization this paper targets is the reduction of the number and/or granularity of features collected for analysis. Features can either be completely suppressed (removed) or generalized. Generalization involves replacing a value with a less specific but semantically consistent value. For example, instead of an exact age, represented by the domain of integers between 0 and 120, a generalized age may consist of 10-year ranges. 

The generalization techniques we employ are similar to those used to achieve k-anonymity on datasets, a process called \emph{anonymization} \cite{Sweeney02}. Although we do not aim to release datasets, only to generalize newly collected data for analysis. Moreover, generalizations enacted for anonymization typically try to minimize data loss while still achieving k-anonymity. In our approach, generalizations are targeted at \textbf{minimizing model accuracy loss}, while incurring the \textbf{largest possible data loss} to increase privacy protection. This is done by taking into account knowledge of the ML algorithm and optimizing the generalizations to minimize their harmful effect on the accuracy of the model. 
An iterative process continues to generalize features until it reaches an accuracy threshold, beyond which we are not willing to compromise for the goal of data minimization.  Once the generalized feature set is determined, any new data collected for analysis can be generalized while it is being collected, before it is fed into the model, thus achieving data minimization. 

To the best of our knowledge, our contribution is a first-of-a-kind method for adhering to the data minimization principle for ML models. Our method enables minimizing the adverse effect of the minimization on the model's accuracy, and does not require any changes to the original model. In addition, we propose an extension to this method that allows dynamic collection and minimization of data in a manner determined by the data subject, providing \emph{personalized data minimization}.

The remainder of the paper is organized as follows. We provide background information in Section \ref{sec:Background}, present the data minimization procedure in Section \ref{sec:dmp}, and offer an evaluation of our method in Section \ref{sec:evaluation}. We discuss some design choices and alternatives in Section \ref{sec:discussion}, present related work in Section \ref{sec:related_work}, and conclude in Section \ref{sec:conclusion}.

\section{Background}
\label{sec:Background}

\subsection{k-anonymity and Generalization}
\label{sub:gen_types}

K-anonymity was proposed by L. Sweeney \cite{Sweeney02} to address the problem of releasing personal data while preserving individual privacy. The approach is based on generalizing attributes and possibly deleting records until each record becomes indistinguishable from at least k-1 other records. 

Most k-anonymity algorithms are based on finding groups of similar records of at least size $k$ that can be generalized together to a single value (on the quasi-identifiers), thus fulfilling the indistinguishability requirement.
A generalization of a numeric attribute typically consists of a range of consecutive values, whereas the generalization of a categorical attribute consists of a sub-group of categories. The example below shows two clusters of similar records have been identified, each with a specific generalization of two features. 

\begin{center}
	\begin{lstlisting}[basicstyle=\tiny]
	{
	   "clusterA": {
	      "featureA": { "start": 0.0, "end": 1.2 },
	      "featureB": { "categories": ["a", "b", "d"] }
	   },
	   "clusterB": {
	      "featureA": { "start": 1.3, "end": 4.0 },
	      "featureB": { "categories": ["c"] }
	   }
	}
	\end{lstlisting}
\end{center}

There are essentially two types of generalizations, emph{global recoding} and \emph{local recoding}. Global recoding means that the ranges for each feature are determined once for the entire dataset, i.e., a particular detailed value must be mapped to the same generalized value in all records. In this case we ``force'' the same generalizations for the entire domain, which may result in smaller ranges. 
Local recoding, allows the same detailed value to be mapped to different generalized values in each group. Having different ranges for different areas in the domain usually enables better generalizations. 

\subsection{Information Loss Metrics}
\label{sub:metrics}
Many metrics for measuring information loss have been developed. A recent 2018 survey noted 78 distinct metrics, while grouping them into different categories and intended uses \cite{Wagner18}. 
The Classification Metric (CM) \cite{Iyengar02} is suitable when the purpose of the anonymized data is to train a classifier. Each record is assigned a class label, and information
loss is computed based on the adherence of a tuple to the majority class of its group.
The Discernibility Metric (DM) \cite{Bayardo05} measures the cardinality of the equivalence class.

The Generalized Loss Metric \cite{Iyengar02} and the similar Normalized Certainty Penalty (NCP) \cite{Ghinita07}, \cite{Xu06} are considered more accurate.
The latter has the advantage of being able to measure information loss across different datasets.
For a numerical attribute with domain $D$, NCP of a generalized range $G$ is defined as:
\begin{equation}
 NCP_{A\ num}(G) = \frac{max(G)-min(G)}{max(D)-min(D)} 
 \end{equation}
Similarly, for categorical attributes:
\begin{equation}
 NCP_{A\ cat}(G) =
\begin{cases}
0               & no\ generalization \\
\frac{|values(G)|}{|values(D)|} & otherwise
\end{cases}
\end{equation}
where $|values(G)|$ represents the number of distinct values in the generalization and
$|values(D)|$ represents the number of distinct values in the domain.
The NCP over all generalized attributes is computed using a weighted average.
Finally, the Global Certainty Penalty (GCP) is computed over all instances in a dataset (for more details see \cite{Xu06}).

\section{Data Minimization Process}
\label{sec:dmp}
The process starts with an existing machine learning model and a dataset consisting of records along with the model's predictions for them. The existing model's predictions are used as the labels for our learning process. We later refer to this labeled dataset as the training dataset, however it is important to note that it does not necessarily have to be the same dataset that was used to train the model. The main reason for this decision is that the original labeled training data may no longer be available, whereas generating predictions for some dataset given an existing model is simple. Moreover, the original model may have learned some very complex structures, but learning its ``decision boundaries'' might be easier. If the original training data is available, it can of course be used.

Our goal is to use the model's predictions to guide the creation of groups of similar records as a basis for the generalization process; in this way, we get a generalization that is tailored to the model. Throughout the remainder of the paper, we refer to these groups as clusters.
The desired accuracy must also be supplied. This may be equal to the original model's accuracy, if no degradation is allowed, or a percentage of deviation from the original accuracy. 

The result of the process is a generalization of the input features, as exemplified in Section \ref{sub:gen_types}. Some features may be completely suppressed, and others may be generalized. The process does not involve retraining the original model or making any changes to it; the generalizations are only applied to newly collected data for analysis, i.e., runtime data. Thus, it is particularly suited for existing systems.
Our basic implementation yields a global recoding of the data. Later in this paper we discuss how other types of generalizations can be achieved. 

As presented in Section \ref{sub:metrics}, there are several different quality metrics that can be used to measure the degree of generalization or privacy of a dataset. 
We chose to use the NCP metric (\cite{Ghinita07}) to measure the quality of the resulting generalization. This metric basically looks at the sizes of the generalized ranges relative to the original feature domain, averaging this score across all records in the dataset. Since our goal is to maximize privacy preservation within the given accuracy constraints, in our case high information loss (NCP) is a desired quality. We did not use weights for the different features; in Section \ref{sec:discussion}, we elaborate on how different feature weights may be used.

\subsection{Generalizer Training}
\label{sub:gen}
We start by \textbf{training a generalizer model} on the training dataset (labeled with the original model's predictions), i.e., we train a model to predict the original model's predictions. The goal of this model is to learn the ``decision boundaries'' of the original model. 
We chose to use a univariate decision tree as the generalizer model, since the splits that the tree creates on each internal node can be used as a basis to determine the generalized ranges. We use the leaf nodes of the tree as our groups of similar inputs and create the generalizations based on the decisions on the tree path leading to each leaf. (Features that are not needed to arrive at a specific leaf may receive any value.) 
Note that a univariate decision tree will always create boundaries that are straight lines, parallel to the axes. In Section \ref{sec:discussion}, we discuss other generalizer models that can be used instead of a decision tree, especially for cases where a decision tree may not be powerful enough to account for the complexity of the original model.

Since the goal is to find the best generalization without harming the model's accuracy, we start by generating a decision tree with homogeneous leaves. Each leaf contains only inputs that generate the same prediction in the original model. We then \textbf{derive the initial set of generalizations} by combining all split values of each feature from the tree's internal nodes.

After the initial set of generalized features is obtained, we \textbf{apply the generalizations} to the test data (we explain how later in this section) and check the accuracy of the original model on it. We \textbf{measure the relative accuracy}, i.e., what percentage of the original predictions are retained when applying the model to the generalized data. Based on the relative accuracy measured, a decision is made on whether to continue the process or not. If the accuracy threshold is reached, the generalizations derived directly from the generalizer model are used. If the accuracy is lower or higher than the desired threshold, we proceed to perform additional steps to improve either the accuracy or the generalization. At the end of each such iteration, the resulting accuracy is again measured by applying the current generalization to the test data and checking the accuracy.

If the achieved accuracy is higher than the threshold, we employ a step to \textbf{improve the generalization}. This is done by iteratively pruning the decision tree, i.e., going up from the leaves to higher nodes in the tree. Each such pruning effectively removes (at least) one split value for one of the features, combining two lower level ranges into a single range and reducing the overall number of ranges for that feature. 
In our implementation we did not use any sophisticated means to choose which nodes to prune. The algorithm simply goes up one level in the entire tree simultaneously. We continue to rise in the tree, one level at a time, until we reach the root node or until the accuracy threshold is met.

If the achieved accuracy is lower than the threshold, we employ a step designed to \textbf{improve the accuracy} by removing features from the generalization. This means that instead of generalizing it, this feature will be left unchanged. 

For this step, we define an additional metric called ILAG, inspired by the score function from \cite{Fung05}. In their case, the authors' goal was to maximize the information gain for each unit of anonymity loss. We adapted this to our setting, where we seek to maximize the information loss (NCP score) for each unit of accuracy gain. We define ILAG for a feature $f$ as:
\begin{equation}
	ILAG(f)=
	\begin{cases}
		\frac{NCP(f)}{AccuracyGain(f)}, & \text{if}\ AccuracyGain(f)\ne0 \\
		NCP(f), & \text{otherwise}
	\end{cases}
\end{equation}

This is the measure we seek to maximize in our algorithm. In each iteration, we choose the feature with the lowest ILAG score and remove it from the generalization. This step repeats until all features are removed or until the accuracy threshold is met. 

To check the original model's accuracy on generalized data without making changes to the model, each test sample must be mapped to a concrete, representative point in the domain of the original features. The idea is to map all records belonging to the same cluster (leaf node) to the same representative value. We thus achieve generalization by mapping multiple different original values to the same concrete value. 
The choice of representatives may greatly affect the method's performance. As the representative value, we chose to use an actual point closest to the median of the cluster, that the model classifies as the majority class for that cluster. 

\begin{figure}
	\centering
	\includegraphics[width=10cm]{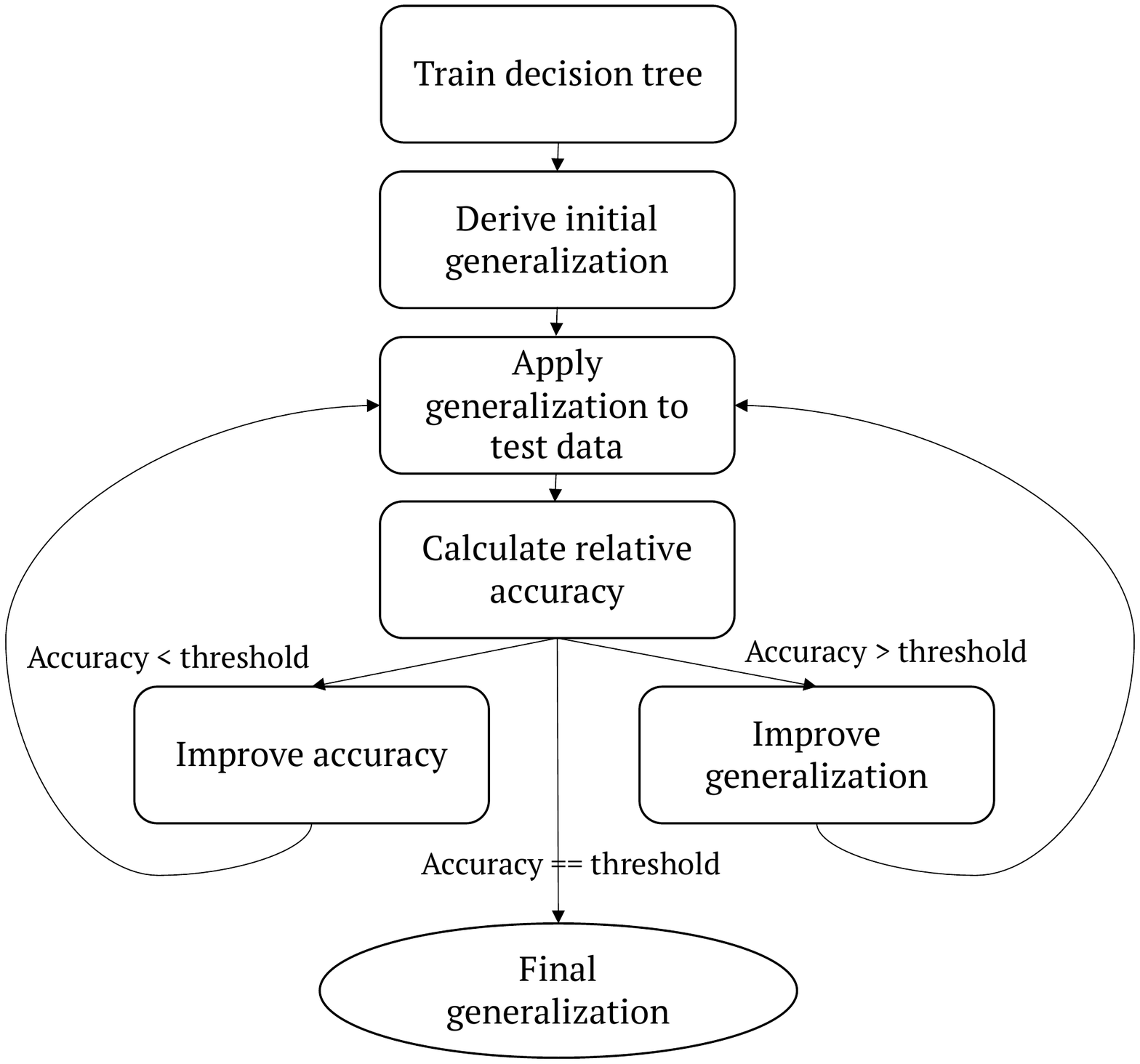}
	\caption{Complete minimization process}
	\label{fig:process}
\end{figure}

The minimization process eventually \textbf{yields a minimal dataset} required to achieve the required level of accuracy. The output is a set of generalized feature ranges tailored to a specific ML model. This set of generalized features can be used whenever collecting new data for analysis. The complete minimization process is presented in Figure \ref{fig:process}.

This process is typically performed after applying ``regular'' feature selection that chooses the most influential features for the model. This process may use any known feature selection techniques, as would be normally applied for that type of model. Our method may also be applied without any prior feature selection for models in which it is not typically applied (such as neural networks).
%
%If the original model is retrained some time after the generalization process was performed, it is recommended to retrain the generalizer as well and derive new generalizations that are best suited for the new model. It is of course possible to continue using the same generalizations, however the model's accuracy may be affected. It is also possible to use newly collected and generalized data to retrain the model (e.g., for reinforcement learning), however, in this case as well, the model's accuracy may be further degraded.

\subsection{Applying Generalizations to Newly Collected Data}

Once the generalized feature set is determined, there are several ways to collect new data for classification. The first option is to use the feature ranges computed as described in the previous section (by combining all split values for each feature). This results in a global recoding of the data, i.e., each feature has a predetermined set of ranges. When using this approach, the user whose data is collected never actually divulges their exact data, only the relevant ranges. 

A second option is to have a \emph{minimization procedure}, in which a piece of code maps original data points to generalized data points, depending on the cluster to which the data point belongs. This piece of code can run at the endpoint where data is collected and immediately generalize the raw data before sending it to analysis; for example, on a server or cloud. Such a minimization procedure can result in a local recoding, thus potentially enabling better generalizations.

The advantage of the first option is that the minimization is more transparent to users. At no point do they disclose their exact information, so they can be sure it is never used. However, the level of generalization achieved (for the same accuracy level) will be lower than when employing the second option. 

Another option is to dynamically determine the ranges presented to the user, based on their choices for previous features. Each time a generalized value is selected, this information can be used to dynamically improve the generalizations for other features. This is possible because a feature's generalizations may have been restricted by domains that are no longer relevant once the value of another feature is known.
The order of filling in the feature values may be determined by the user. This allows them to reduce the granularity, or even avoid disclosing altogether, the information that they consider most sensitive, achieving \emph{personalized data minimization}. Another alternative is to allow the organization collecting the data to decide the order of features, for example based on cost.  

%In Section \ref{sec:pdm} we present an advanced approach that dynamically determines the ranges presented to the user.

Note that the NCP scores presented in Section \ref{sub:results} were computed assuming the global recoding generated by the basic algorithm. Applying the minimization procedure or dynamic minimization would likely improve the information loss. 

\section{Evaluation}
\label{sec:evaluation}
\subsection{Methodology}
Our evaluation method consists of the following steps. First, we select a dataset and train one or more original models on it (first applying feature selection). We consider the resulting model and its accuracy as our baseline. We then perform the data minimization process described in Section \ref{sec:dmp} and apply the resulting generalizations to a validation set. Finally, we measure the accuracy of the original model on the generalized data. In addition, we compute the information loss achieved by the resulting generalization as a measure for how well the data was generalized: the higher the information loss, the better for our purpose.

We evaluated our method using four openly available datasets: the Adult dataset\footnote{https://archive.ics.uci.edu/ml/datasets/adult}, an excerpt of the 1994 U.S. Census database, which is widely used to evaluate anonymization and generalization techniques; the Nursery dataset\footnote{https://archive.ics.uci.edu/ml/datasets/nursery}, derived from a hierarchical decision model developed to rank applications for nursery schools in Slovenia, which was used by Bild et al. \cite{Bild18} to compare approaches for differentially private statistical classification; the GSS marital happiness dataset\footnote{http://byuresearch.org/ssrp/downloads/GSShappiness.pdf}, a subset of the GSS data created by Joseph Price to study various societal effects of pornography, used by Fredrikson et al. \cite{Fredrikson15} to evaluate the effectiveness of a model inversion attack. In addition, to show that our approach works well for data with a larger number of attributes, we used the Loan dataset\footnote{https://www.lendingclub.com/info/download-data.action}, an excerpt of the Lending Club loan data from 2015.

A summary of the characteristics of each dataset is presented in Table \ref{table:data}. For each dataset the table shows: the number of records in the downloaded data, the number of records actually used, the number of features in the downloaded data, the number of features used (before applying feature selection as part of the training process), the number of categorical features, and the label feature. Note that we removed records that did not have a value for the label feature (which was the case for almost half of the records in the GSS dataset) or whose label value was too scarce. We also removed non-numeric features with too many distinct values (e.g., free-text), features with more than 50\% missing data and features with too high a correlation with the label. For each dataset we trained three types of classifiers: Random Forest, XGBoost, and Neural Network (with 1 hidden layer and 100 neurons). The classifiers' baseline accuracy for each of the datasets is also brough in Table \ref{table:data}.

\begin{table}
	\centering
	\begin{tabular}{p{1.2cm}||p{1cm}|p{1cm}|p{0.8cm}|p{0.8cm}|p{0.5cm}|p{1.8cm}||p{1.3cm}|p{1.5cm}|p{1.3cm}}
		Name & total rows & rows used & total attrs & attrs used & cat & label & Random Forest & XGBoost & Neural Network\\
		\hline \hline
		Adult & 48842 &  48842 & 14 & 12 & 7 & income & 88.08 &  85.61 & 84.56\\
		\hline
		Nursery & 12960 & 12958 &  8 & 8 & 7 & nursery & 97.37 & 97.84 &  98.61\\
		\hline
		GSS & 51020 & 24455 & 8 & 8 & 6 & marital happiness & 62.34 & 68.11 & 67.49\\
		\hline
		Loan & 421095 &  421095 & 144 & 43 & 11 & loan status & 97.8 &  92.91 & 93.48\\
	\end{tabular}
	\caption{Summary of datasets used for evaluation}
	\label{table:data}
\end{table}

 Each dataset was divided into four separate subsets, used to: (1) train the original model, (2) train our generalizer decision tree model, (3) optimize the resulting generalization (choosing what level in the tree and which features to use in the generalization), and (4) final validation. We used a separate dataset to train our generalizer decision tree model because the original training set may not be available. Finally, validation was performed by measuring the relative accuracy of the classifier on the generalized data, i.e., what percentage of the original predictions are retained when applying it to the generalized data.

%\begin{table}
%	\centering
%	\begin{tabular}{p{1.5cm}||p{2.5cm}|p{2.5cm}|p{2.5cm}}
%		Name & Random Forest & XGBoost & Neural Network \\
%		\hline \hline
%		Adult & 88.08 &  85.61 & 84.56 \\
%		\hline
%		Nursery & 97.37 & 97.84 &  98.61\\
%		\hline
%		GSS & 62.34 & 68.11 & 67.49 \\
%		\hline
%		Loan & 97.8 &  92.91 & 93.48 \\
%	\end{tabular}
%	\caption{Original accuracy of different models and datasets}
%	\label{table:originalAccuracy}
%\end{table}

\subsection{Results}
\label{sub:results}
In this section, we present the results of applying our minimization process to the above-mentioned datasets and models. The code used to perform data minimization is available on GitHub\footnote{https://github.com/IBM/ai-minimization-toolkit}. Each graph presents the accuracy on the generalized data as a function of the achieved NCP score. NCP values are between 0 and 1, and represent the degree of information loss: 0 signifying that no information was lost, i.e., preserving the original data, and 1 signifying that all information was lost, i.e., all samples were mapped to the same representative value. Accuracy was measured as a percentage relative to the accuracy of the baseline model. 100\% signifies that all predictions made on the generalized data were identical to those made on the original data. The reason for using the original model's accuracy as the baseline is that we do not aim to improve on the predicitions of the original model, only to get as close to them as possible. In addition, true labels may not be available.

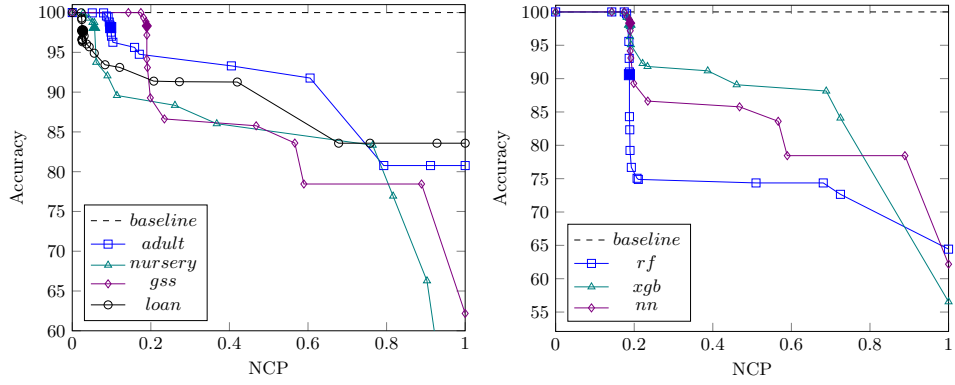
\begin{figure*}
	\hspace*{0.2in}
	\begin{subfigure}{.5\textwidth}
		\resizebox{.88\columnwidth}{!}{%
		\begin{tikzpicture}[baseline,trim axis left]
		\begin{axis}[legend style={font=\small}, title style={font=\small}, xlabel=NCP, ylabel=Accuracy, xmin=0,xmax=1, ymin=60, ymax=101, ytick={60, 65, 70, 75, 80, 85, 90, 95, 100}, 
		legend pos=south west, at={(-1,0)},anchor=west, y label style={at={(-0.1,0.5)},anchor=south}]
		\addplot [mark=none, black, dashed] coordinates { (0,100) (1,100) }; 
		\addlegendentry{$baseline$}
		\addplot [color=blue, mark=square] coordinates {( 0, 100 )	( 0.050989, 99.96 )	( 0.079766, 99.96 )	( 0.08701, 99.57 )	( 0.091302, 99.447 ) ( 0.094215, 98.81 ) ( 0.095658, 98.42 ) ( 0.096918, 98.28 ) ( 0.097896, 98.11 )  ( 0.098226, 98.05 )  ( 0.098836, 97.54 )  ( 0.099899, 97.03 )  ( 0.103423, 96.27 )  ( 0.158592, 95.62 )  ( 0.170705, 94.76 )  ( 0.404988, 93.3 )  ( 0.605282, 91.77 )  ( 0.792809, 80.77 )  ( 0.911821, 80.77 ) (1, 80.77)}; 
		\addlegendentry{$adult$}
		\addplot [color=teal, mark=triangle] coordinates {( 0, 100 )	( 0.02736, 99.92 )	( 0.041647, 99.3 )	( 0.051436, 98.76 )	( 0.05623, 98.76 ) ( 0.060992, 93.75 ) ( 0.089574, 92.05 ) ( 0.113747, 89.58 ) ( 0.261429, 88.34 )  ( 0.367791, 86.03 )  ( 0.764738, 83.33)  ( 0.81659, 76.93 )  ( 0.902778, 66.28 )  ( 1, 33.33 )}; 
		\addlegendentry{$nursery$}
		\addplot [color=violet, mark=diamond] coordinates {( 0, 100 )	( 0.142857, 100 )	( 0.174658, 100 )	( 0.181665, 99.34 )	( 0.187603, 98.85 ) ( 0.188834, 98.36 ) ( 0.18968, 98.32 ) ( 0.189703, 98.29 ) ( 0.189889, 97.17 )  ( 0.189749, 94.15 )  ( 0.191345, 93.09)  ( 0.198804, 89.28 )  ( 0.234521, 86.63 )  ( 0.468101, 85.77 )  ( 0.566425, 83.6 )  ( 0.589488, 78.45 ) ( 0.888944, 78.45 ) ( 1, 62.18 )}; 
		\addlegendentry{$gss$}
		\addplot [color=black, mark=o] coordinates {( 0, 100 )	(0.022727, 99.97 )	( 0.023821, 99.31 )	( 0.024317, 99.12 )	( 0.024733, 96.66 ) ( 0.025082, 96.57 ) ( 0.025399, 96.51 ) ( 0.02561, 96.54 ) ( 0.025798, 96.33 ) ( 0.026151, 97.61 ) ( 0.02628, 97.73 ) ( 0.026481, 97.73) ( 0.02694, 97.66 ) ( 0.027999, 97.47 ) ( 0.030209, 97.06 ) ( 0.036535, 96.05 ) ( 0.043108, 95.7 ) ( 0.056037, 94.89 ) ( 0.083404, 93.44 ) ( 0.120684, 93.11 ) ( 0.207432, 91.38 ) ( 0.272766, 91.31 ) ( 0.419939, 91.27 )  ( 0.677926, 83.58 )   ( 0.757836, 83.58 )   ( 0.928253, 83.58 ) (1, 83.58) };
		\addlegendentry{$loan$}
		\addplot [color=blue, mark=square*, mark color=blue, mark size = 2.5] coordinates {(0.097896, 98.11)}; 
		\addplot [color=violet, mark=diamond*, mark color=violet, mark size = 3] coordinates {(0.18968, 98.32)}; 
		\addplot [color=black, mark=oplus*, mark color=black, mark size = 2.5] coordinates {(0.02628, 97.73)}; 
		\addplot [color=teal, mark=triangle*, mark color=teal, mark size = 3] coordinates {(0.05623, 98.14)}; 
		\end{axis}
		\end{tikzpicture}}
		\label{fig:nn_graph}
	\end{subfigure}%
	\hspace*{0.1in}
	\begin{subfigure}{.5\textwidth}
		\resizebox{.88\columnwidth}{!}{%
		\begin{tikzpicture}[baseline,trim axis left]
		\begin{axis}[legend style={font=\small}, title style={font=\small}, xlabel=NCP, ylabel=Accuracy, xmin=0,xmax=1, ymax=101, ytick={55, 60, 65, 70, 75, 80, 85, 90, 95, 100},  
		legend pos=south west, at={(-1,0)},anchor=west, y label style={at={(-0.1,0.5)},anchor=south}]
		\addplot [mark=none, black, dashed, samples=2] coordinates {( 0, 100 ) (1, 100)}; 
		\addlegendentry{$baseline$}
		\addplot [color=blue, mark=square] coordinates {(0,100) (0.142857, 100) (0.174658, 100) (0.180596, 99.71) (0.186358, 95.58) (0.187205, 93.04) (0.187778, 91) (0.187803, 84.3) (0.188665, 82.33) (0.189291, 79.23) (0.192675, 76.69) (0.207238, 75.06) (0.210427, 74.89) (0.509997, 74.36) (0.68111, 74.36) (0.724863, 72.65)  (1, 64.43)};
		\addlegendentry{$rf$}
		\addplot [color=teal, mark=triangle] coordinates { ( 0, 100 )	( 0.14285, 100 )	( 0.174658, 100 )	( 0.180731, 99.18 )	( 0.186669, 99.06 ) ( 0.187841, 98.12 ) ( 0.188688, 98.16 ) ( 0.188708, 97.79 ) ( 0.189368, 96.28 )  ( 0.191452, 95.09 ) ( 0.221412, 92.27 ) ( 0.233774, 91.82 ) ( 0.387093, 91.17 ) ( 0.460713, 89.08 ) ( 0.688688, 88.14 ) ( 0.724863, 84.09 ) ( 1, 56.54 )}; 
		\addlegendentry{$xgb$}
		\addplot [color=violet, mark=diamond] coordinates {	( 0, 100 )	( 0.142857, 100 )	( 0.174658, 100 )	( 0.181665, 99.34 )	( 0.187603, 98.85 ) ( 0.188834, 98.36 ) ( 0.18968, 98.32 ) ( 0.189703, 98.29 ) ( 0.189889, 97.17 )  ( 0.189749, 94.15 )  ( 0.191345, 93.09)  ( 0.198804, 89.28 )  ( 0.234521, 86.63 )  ( 0.468101, 85.77 )  ( 0.566425, 83.6 )  ( 0.589488, 78.45 ) ( 0.888944, 78.45 ) ( 1, 62.18 )}; 
		\addlegendentry{$nn$}
		\addplot [color=blue, mark=square*, mark color=blue, mark size = 2.5] coordinates {(0.187778, 90.59)}; 
		\addplot [color=teal, mark=triangle*, mark color=teal, mark size = 3] coordinates {(0.188688, 98.16)}; 
		\addplot [color=violet, mark=diamond*, mark color=violet, mark size = 3] coordinates {(0.18968, 98.32)}; 
		\end{axis}
		\label{graph:gss}
		\end{tikzpicture}}
		\label{fig:gss_graph}
	\end{subfigure}%
	\caption{Results for different models and datasets: (a) Accuracy vs. NCP for different datasets using Neural Network model (b) Accuracy vs. NCP for different models on GSS data}
	\label{fig:graph}
\end{figure*}

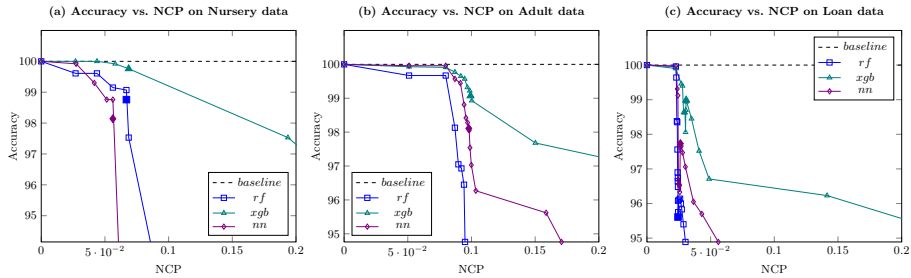
\begin{figure*}
	\hspace*{0.2in}
	\begin{subfigure}{.33\textwidth}
		\resizebox{.88\columnwidth}{!}{%
			\begin{tikzpicture}[baseline,trim axis left]
			\begin{axis}[legend style={font=\small}, title style={font=\small}, xlabel=NCP, ylabel=Accuracy, xmin=0,xmax=0.2, ymin=94.15, ymax=101, ytick={95, 96, 97, 98, 99, 100}, title={\bfseries (a) Accuracy vs. NCP on Nursery data}, 
			legend pos=south east, at={(-1,0)},anchor=west, y label style={at={(-0.07,0.5)},anchor=south}]
			\addplot [mark=none, black, dashed, samples=2] coordinates {( 0, 100 ) (0.2, 100)}; 
			\addlegendentry{$baseline$}
			\addplot [color=blue, mark=square] coordinates {(0,100) (0.026942, 99.61) (0.043724, 99.61) (0.056247, 99.15) (0.066856, 99.07) (0.068653, 97.53)  (0.094031, 92.51)};
			\addlegendentry{$rf$}
			\addplot [color=teal, mark=triangle] coordinates {( 0, 100 )	( 0.026942, 100 )	( 0.043724, 100 )	( 0.058043, 99.92 )	( 0.068653, 99.76 ) ( 0.193653, 97.53 ) ( 0.312816, 93.44 )}; 
			\addlegendentry{$xgb$}
			\addplot [color=violet, mark=diamond] coordinates {	( 0, 100 ) ( 0.02736, 99.92 )	( 0.041647, 99.3 )	( 0.051436, 98.76 )	( 0.05623, 98.76 ) ( 0.060992, 93.75 ) }; 
			\addlegendentry{$nn$}
			\addplot [color=blue, mark=square*, mark color=blue, mark size = 2.5] coordinates {(0.066856, 98.76)}; 
			\addplot [color=teal, mark=triangle*, mark color=teal, mark size = 3] coordinates {(0.068653, 99.76)}; 
			\addplot [color=violet, mark=diamond*, mark color=violet, mark size = 3] coordinates {(0.05623, 98.14)}; 
			\end{axis}
			\label{graph:nursery_zoom}
			\end{tikzpicture}%
		}
	\end{subfigure}%
	\begin{subfigure}{.33\textwidth}
		\resizebox{.88\columnwidth}{!}{%
			\begin{tikzpicture}[baseline,trim axis left]
			\begin{axis}[legend style={font=\small}, title style={font=\small}, xlabel=NCP, ylabel=Accuracy, xmin=0,xmax=0.2, ymin=94.76, ymax=101, ytick={95, 96, 97, 98, 99, 100}, title={\bfseries (b) Accuracy vs. NCP on Adult data}, 
			legend pos=south west, at={(-1,0)},anchor=west, y label style={at={(-0.07,0.5)},anchor=south}]
			\addplot [mark=none, black, dashed, samples=2] coordinates {( 0, 100 ) (0.2, 100)}; 
			\addlegendentry{$baseline$}
			\addplot [color=blue, mark=square] coordinates {(0,100) (0.050989, 99.67) (0.079766, 99.67) (0.08701, 98.13) (0.089742, 97.05) (0.092118, 96.93) (0.094141, 96.45) (0.094946, 94.76) };
			\addlegendentry{$rf$}
			\addplot [color=teal, mark=triangle] coordinates { ( 0, 100 )	( 0.050989, 99.93 )	( 0.079766, 99.91 )	( 0.08701, 99.77 )	( 0.091673, 99.65 ) ( 0.094711, 99.57 ) ( 0.096808, 99.32 ) ( 0.098658, 99.22 ) ( 0.099189, 99.09 )  ( 0.099845, 99.07 ) ( 0.100211, 98.93 ) ( 0.150225, 97.68 ) ( 0.229902, 97.03 ) ( 0.240841, 96.56 ) ( 0.2919, 96.35 ) ( 0.403391, 95.08 ) }; 
			\addlegendentry{$xgb$}
			\addplot [color=violet, mark=diamond] coordinates {	( 0, 100 )	( 0.050989, 99.96 )	( 0.079766, 99.96 )	( 0.08701, 99.57 )	( 0.091302, 99.447 ) ( 0.094215, 98.81 ) ( 0.095658, 98.42 ) ( 0.096918, 98.28 ) ( 0.097896, 98.11 )  ( 0.098226, 98.05 )  ( 0.098836, 97.54 )  ( 0.099899, 97.03 )  ( 0.103423, 96.27 )  ( 0.158592, 95.62 ) ( 0.170705, 94.76 ) }; 
			\addlegendentry{$nn$}
			\addplot [color=blue, mark=square*, mark color=blue, mark size = 2.5] coordinates {(0.095604, 92.77)}; 
			\addplot [color=teal, mark=triangle*, mark color=teal, mark size = 3] coordinates {(0.099189, 99.07)}; 
			\addplot [color=violet, mark=diamond*, mark color=violet, mark size = 3] coordinates {(0.097896, 98.11)}; 
			\end{axis}
			\label{graph:adult_zoom}
			\end{tikzpicture}%
		}
	\end{subfigure}%
	\begin{subfigure}{.33\textwidth}
		\resizebox{.88\columnwidth}{!}{%
			\begin{tikzpicture}[baseline,trim axis left]
			\begin{axis}[legend style={font=\small}, title style={font=\small}, xlabel=NCP, ylabel=Accuracy, xmin=0,xmax=0.2, ymin=94.89, ymax=101, ytick={95, 96, 97, 98, 99, 100}, title={\bfseries (c) Accuracy vs. NCP on Loan data}, 
			legend pos=north east, at={(-1,0)},anchor=west, y label style={at={(-0.07,0.5)},anchor=south}]
			\addplot [mark=none, black, dashed, samples=2] coordinates {( 0, 100 ) (0.2, 100)}; 
			\addlegendentry{$baseline$}
			\addplot [color=blue, mark=square] coordinates {(0,100) (0.022727, 99.96) (0.023099, 99.64) (0.023485, 98.38) (0.023838, 98.35)  (0.023972, 97.56)  (0.024103, 96.9)  (0.024219, 96.65)  (0.02432, 96.74)  (0.02438, 96.48)  (0.024397, 96.08)  (0.024402, 95.61)  (0.024501, 95.75)  (0.025054, 96.11)  (0.02604, 96.14)  (0.026698, 96)  (0.02736, 95.83) (0.028698, 95.4)  (0.030256, 94.89) };
			\addlegendentry{$rf$}
			\addplot [color=teal, mark=triangle] coordinates { ( 0, 100 )	( 0.022727, 99.9 )	( 0.026889, 99.4600 )	( 0.027923, 99.4 )	( 0.028752, 98.63 ) ( 0.029508, 98.63 ) ( 0.029921, 98.63 ) ( 0.030319, 98.06 ) ( 0.030517, 98.7 )  (0.030705, 98.95 )  (0.030776, 98.95 )  (0.030814, 98.99 )  (0.031103, 98.95 )  (0.035003, 98.45 )  (0.040935, 97.52 )  (0.048802, 96.71 )  (0.141447, 96.23 )  (0.258826, 94.9 ) }; 
			\addlegendentry{$xgb$}
			\addplot [color=violet, mark=diamond] coordinates {	( 0, 100 )	(0.022727, 99.97 )	( 0.023821, 99.31 )	( 0.024317, 99.12 )	( 0.024733, 96.66 ) ( 0.025082, 96.57 ) ( 0.025399, 96.51 ) ( 0.02561, 96.54 ) ( 0.025798, 96.33 ) ( 0.026151, 97.61 ) ( 0.02628, 97.73 ) ( 0.026481, 97.73) ( 0.02694, 97.66 ) ( 0.027999, 97.47 ) ( 0.030209, 97.06 ) ( 0.036535, 96.05 ) ( 0.043108, 95.7 ) ( 0.056037, 94.89 ) }; 
			\addlegendentry{$nn$}
			\addplot [color=blue, mark=square*, mark color=blue, mark size = 2.5] coordinates {(0.024402, 95.61)}; 
			\addplot [color=teal, mark=triangle*, mark color=teal, mark size = 3] coordinates {(0.030814, 98.99)}; 
			\addplot [color=violet, mark=diamond*, mark color=violet, mark size = 3] coordinates {(0.02628, 97.73)}; 
			\end{axis}
			\label{graph:loan_zoom}
			\end{tikzpicture}%
		}
	\end{subfigure}
	\caption{Results for Nursery, Adult and Loan datasets}
	\label{fig:graphs}
\end{figure*}

Figure \ref{fig:graph}.a presents a comparison of the results for the different datasets when using a neural network model. The remainder of the graphs show the results for a single dataset, comparing the different tested models: random forest (rf), XGBoost (xgb) and neural network (nn). Figure \ref{fig:graph}.b presents the results for the GSS dataset and Figure \ref{fig:graphs} presents the results for the other datasets.

Next we present a few examples of generalizations that were achieved. We show the generalizations both for the case of no accuracy loss at all, as well as the case where a 2\% relative accuracy loss is acceptable. Table \ref{table:gss_example} presents the achieved generalizations for the GSS dataset, using a neural network model. 
\begin{table*}
	\centering
	\begin{tabular}{p{2cm}|p{7.5cm}|p{2.5cm}}
		Feature & 2\% relative accuracy loss & No accuracy loss \\
		\hline \hline
		Marital status & \textcolor{blue}{Not needed} & \textcolor{blue}{Not needed} \\
		\hline
		Happiness & [Pretty happy, Not too happy, Other], [Very happy]  & Same as 2\%	 \\
		\hline
		Race & [Black, Other], [White] & \textcolor{red}{Not generalized} \\
		\hline
		Work status & [Temp not working, Unemployed - laid off, School, Other], [Working fulltime], [Keeping house], [Retired], [Working parttime] & \textcolor{red}{Not generalized} \\
		\hline
		Age & 54 ranges representing values 0-89 & \textcolor{red}{Not generalized} \\
		\hline
		Children & \textcolor{red}{Not generalized} & \textcolor{red}{Not generalized} \\
		\hline		
		X rated & \textcolor{red}{Not generalized} & \textcolor{red}{Not generalized} \\
		\hline  \hline
		NCP value: & \emph{0.189703} & \emph{0.174658} \\
	\end{tabular}
	\caption{Example of generalizations for the GSS dataset}
	\label{table:gss_example}	
\end{table*}

Several aspects of these results are worth noting. First, even when reaching an NCP score of 1, where all features are generalized to the entire domain of the feature, we do not see a drop of the model accuracy below 33\%, in some cases 80\%. This ``null accuracy'' represents the accuracy that can be achieved by always predicting the most frequent class. Similarly, we are always able to reach 100\% relative accuracy. This is because, in the worst case, we revert back to the original data. It is important to note that cases exists (e.g., with the GSS data) where we are able to reach 100\% even with some generalization. This illustrates the capability of our approach to identify information that is not necessary and generalize it without impacting accuracy.

Second, we can see that this method performs best for XGBoost models and worst for random forest models in most of the tested datasets, regardless of the original accuracy of each model. We believe this is related to the size and complexity of the generated decision tree. Typically, the decision tree generated for the XGBoost model is much smaller and has fewer branches compared to the other models. In general, any ML model that is too complex tends to overfit the training data and end up being less accurate. When concentrating on the most relevant area of up to 2\% relative accuracy loss, all models perform similarly, resulting in close NCP values (for each given dataset).

Third, there is also a noticeable difference between datasets, with the GSS dataset achieving the best results and the Loan dataset the worst results. With the GSS dataset, we were able to achieve identical accuracy to the original data while yielding an NCP score of 0.17, or approximately 1/6, in all tested models.
One of the reasons that may lead to a higher NCP score could be the prevalence of categorical features, especially those with a small number of categories. Any generalization of such a feature will automatically have a significant effect on the overall NCP score. Numerical features have a higher degree of freedom, resulting in a large number of ranges, thus yielding a lower NCP score. In addition, in the GSS dataset, one feature was discovered not to be needed at all, which also contributed to its higher NCP score.
%Looking at the resulting generalizations we observed that the datasets with better results usually have several generalized categorical features. 
%In the Loan dataset, only numeric features were generalized, and those have a large number of ranges. 

%The fourth aspect to notice is that in almost all datasets and models, there is a large initial dip in accuracy, that later becomes more moderate. This is because in the initial iterations of tree pruning, there is a relatively small to non-existent increase in the NCP score. This can be explained by either: (1) pruning splits that are redundant, for example in cases where the same split on the same feature also appears in another branch or level of the tree, and thus has no effect on the resulting generalization; or (2) pruning nodes that affect a very small number of samples.

Lastly, we observe in the Loan dataset the accuracy goes up at some point while increasing the NCP score, before going back down. This happens in the phase of the algorithm where features are iteratively removed from the generalization. This may be an indication that in this dataset, for some features, generalizing the feature to a smaller domain actually has a positive effect on the model's accuracy, as suggested by previous works on feature abstraction \cite{Alkabawi17}.

\subsubsection{Effect on disclosure risk}
\label{risk}
Although our main goal is compliance with the data minimization principle in GDPR, we decided to measure how the generalizations we produce affect the disclosure risk of the resulting dataset. Disclosure risk can be defined as the risk that an adversary can use the protected (generalized) dataset to derive confidential information on an individual from the original dataset. This risk would be relevant, for example, if an attacker were able to get hold of the generalized data stored within the organization. Disclosure risk can be divided into two types: identity disclosure, i.e., identifying a specific individual's record in the data, and attribute disclosure, which means inferring sensitive (non-disclosed) information about an individual. Most of the literature deals with the former kind.

Several metrics for measuring disclosure risk have been proposed; we chose the one described in \cite{Kim07}, with a small variation: since we do not have a specific value of $k$, we simply sum the probability of each record's specific combination of quasi-identifier values appearing in the dataset over all records in the dataset. We therefore used the following formula to calculate the risk of a dataset $PT$: 
\begin{equation}
	Risk=\frac{\sum_{r \in PT}^{} \frac{1}{freq(qi(r))}} {\#(r)}
\end{equation}
Where $r$ denotes a single record and $freq(qi(r))$ denotes the frequency of the specific combination of values for the quasi-identifiers in $r$. This risk score takes values between 0 and 1, with 1 coresponding to the case where all records in the dataset are unique.

We measured this risk for three datasets: the two datasets corresponding to the results presented in Table \ref{table:gss_example}, and the original GSS dataset. For simplicity we assumed that all features in the dataset are quasi-identifiers (except for the label). For the original GSS dataset, with no generalizations applied, there were 2085 distinct records out of 2446 total records in the validation set, corresponding to a risk value of 0.788. For the generalized dataset corresponding to no accuracy loss, there are 2054 disctinct records and a risk value of 0.771; only slightly lower than the original data. However, when looking at the dataset that was generated with 2\% accuracy loss, the result is much more remarkable. It contains only 222 disctinct records and has a risk score of 0.047. This is a very significant reduction, of more than an order of magnitude, in the disclosure risk, and demonstrates that our method has value in protecting individuals' privacy in addition to helping companies comply with regulations.

\section{Discussion}
\label{sec:discussion}
In this section we discuss some design choices made in our solution, alternatives that were considered and possible optimizations for the future.

\subsection{Choice of Privacy Metric}
There are many different privacy metrics used for different purposes in the literature. We tried to choose a metric that was suitable for measuring the quality of a generalization. Since generalizations are typically employed in anonymization algorithms, we surveyed several metrics used in previous anonymization-related works and chose the NCP metric to measure the quality of generalizations. 

We also demonstrated in Section \ref{risk} that our proposed process can significantly improve the disclosure risk of the resulting data, thus providing improved privacy to the relevant individuals. It is worth noting that the NCP metric (and ILAG score) can easily be replaced with another privacy or information loss metric, while retaining the overall framework described in Section \ref{sec:dmp}.

NCP also has a weighted variant, presented in \cite{Xu06}, which can be used to assign weights to the different features, thus increasing the effect of certain features over others in the quality measurement. This could be used to assign weights to features based on their sensitivity level and steer the generalization process towards results that prefer better generalizations of the more sensitive features.

\subsection{Alternative Generalization Models}
\label{subsec:models}
In this study we used a univariate decision tree as the generalization model. Decision trees inherently create a global recoding of the feature space, which can be very convenient for presenting ranges to users. However, it is reasonable to assume that some ML models will be difficult to mimic using a simple decision tree, especially in cases of highly complex, non-linear models, such as deep neural networks. It is therefore important to consider alternative generalization models, for example based on clustering.

\subsection{Optimizing the Tree Pruning and Feature Removal Processes}
\label{subsec:opt}
In our implementation, we used the ILAG measure to decide which features to remove from the generalization for cases where the accuracy achieved is below the defined threshold. The phase where the tree is pruned to achieve better generalizations could also be improved by using more sophisticated means to decide which nodes to prune,  either using standard tree pruning techniques \cite{Kearns98} or using the ILAG measure to determine which leaves to prune. 

The sensitivity level of each feature can also be considered when when selecting features for removal or tree nodes for pruning. To this aim, we could employ a weighted variant of NCP that takes this information into account while optimizing the generalization process. 
We could even combine the two directions, pruning and removing features from the generalization, optimizing on both simultaneously to yield better results. Other methods for increasing the accuracy can also be explored, such as splitting one or more ranges instead of completely removing a feature from the generalization. 

\section{Related Work}
\label{sec:related_work}
We are not aware of any existing work that addresses data minimization for ML models. We briefly highlight some works in closely-related areas.

A great body of prior work has focused on protecting the privacy of training sets used to train ML models. Several attacks were shown to be able to reveal either whether an individual was part of the training set (\emph{membership attacks}) or infer certain possibly sensitive properties of the training data (\emph{attribute inference attacks}) \cite{Fredrikson14}, \cite{Fredrikson15}, \cite{Shokri17}, \cite{Salem19}. Therefore, many methods have been proposed to anonymize or de-identify the training data, including  \emph{tailored anonymization} of training data \cite{LeFevre08}, \cite{Iyengar02}, adding noise or applying \emph{differential privacy} to the training process \cite{Zhang18}, \cite{Abadi16}, \cite{Papernot18}. 
However, all of these techniques are aimed at protecting the training set and do not provide any protections for runtime data, i.e., newly collected data to which the model is applied. 

Another focus area has been performing training and/or prediction on encrypted data \cite{Chabanne17}, \cite{Zhu18}, \cite{Barni06}, \cite{Hesamifard18}. These approaches do not solve the issue of collecting less data to begin with. They simply offer better protection for the data that is already collected and stored.

Other works have focused on feature selection and dimensionality reduction, also called feature extraction \cite{Ma10}. These are aimed at fulfilling the \emph{principle of parsimony} by reducing the number of features used as input to the model. However, they do so mainly to improve the performance and memory footprint of the model, not with the goal of privacy in mind. These methods sometimes produce a reduced set of features, but do not reduce the granularity of the collected data. Embedding is another method to achieve dimensionality reduction for high-dimensional data, the most famous being LLE \cite{Roweis00} and word2vec \cite{Mikolov13}.

Feature abstraction \cite{Soleymani18}, \cite{Alkabawi17} is also used for dimensionality reduction. These can sometimes result in different abstraction levels of the original features. Until now, such techniques have been employed with the goal of improving model accuracy and not privacy. Feature discretization \cite{Kotsiantis05}, \cite{Ferreira11} is concerned with the discretization of continuous features.

\section{Conclusion and Future Work}
\label{sec:conclusion}
In this work we presented a novel method to perform data minimization for ML models in a manner that minimizes the effect on model accuracy. To the best of our knowledge, this issue has not been tackled before. Our approach is complementary to standard feature selection and extraction techniques: in addition to suppressing certain features, it can lead to the generalization of other features, thus further reducing the amount and granularity of collected data. This both helps organizations to adhere to the data minimization principle, as well as reduce data collection, storage, and management costs. 

Our method is generic and can be applied to any dataset and any ML model. We have shown that our method was able to achieve good results - finding data to minimize without impacting accuracy - on different types of models.

This is a very initial implementation of data minimization for ML, leaving many areas of possible improvement, some of which are detailed in Section \ref{sec:discussion}. 
\bibliographystyle{splncs04}
\bibliography{minimization}

\end{document}